\documentclass[cameraready]{Interspeech}

\newcommand{\para}[1]{\vspace{0.2cm}\noindent\textbf{#1}\hspace{0.2cm}}

\title{Can Large Language Models Reliably Correct Errors in Low-Resource ASR? A Contamination-Aware Case Study on West Frisian}

\author[]{Yun}{Hao}
\author[]{Reihaneh}{Amooie}
\author[]{Wietse}{de Vries}
\author[]{Rik}{van Noord}
\author[]{Martijn}{Wieling}

\address{
University of Groningen, The Netherlands
  }
\email{\{yun.hao, r.amooie, wietse.de.vries, r.i.k.van.noord, m.b.wieling\}@rug.nl}


\newcommand{\bad}[1]{\textcolor{red}{#1}}
\newcommand{\good}[1]{\textcolor{green!60!black}{#1}}
\usepackage{comment}
\usepackage{multirow}
\usepackage{booktabs}
\usepackage{makecell}
\usepackage{xcolor}
\usepackage{graphicx}
\usepackage{subcaption}

\begin{document}

\maketitle

\begin{abstract}
Automatic speech recognition (ASR) has improved substantially in recent years, yet performance remains limited for low-resource languages. Large language models (LLMs) have shown promise for improving ASR through generative error correction (GER), but their effectiveness in low-resource settings remains underexplored. In addition, it remains unclear to what extent data contamination influences the reported improvements in LLM-based GER.

This study investigates LLM-based GER for low-resource Frisian. In addition to a public corpus, we construct and use a Frisian offline dataset with non-public texts for evaluation to control for potential data contamination.
Results show that GER improves ASR performance in most settings, with the best GPT-5.1 results surpassing oracle WERs. Comparable gains on the offline dataset indicate that improvements reflect true correction ability. We further provide a detailed error analysis revealing model correction patterns.

\vspace{0.1cm}



\end{abstract}

\section{Introduction}

In recent years, automatic speech recognition (ASR) has seen remarkable progress, with substantial gains in recognition accuracy and robustness. Multilingual self-supervised and weakly supervised speech models, including XLS-R \cite{babu22_interspeech} and Whisper \cite{radford2023robust}, have played a central role in improving ASR performance, particularly for low-resource languages. Despite these advances, recognition accuracy remains limited for languages with very scarce transcribed speech data or a small number of speakers \cite{bartelds2023making, yong25_interspeech}.

Conventional methods typically improve ASR performance by combining the acoustic model with a language model (LM), either during the decoding process or by rescoring the N-best ASR hypotheses. Most recently, with the rapid advancement of large language models (LLMs), researchers have begun to explore generative error correction (GER) as a better post-processing strategy for ASR outputs. Rather than selecting among existing hypotheses, GER directly generates revised transcriptions conditioned on ASR outputs, thereby allowing corrections beyond the constraints of the original decoding space. This approach has demonstrated substantial gains in high-resource languages, particularly English \cite{ma2023can, yang2023generative, chen2023hyporadise, yamashita25_interspeech}.
However, it remains unclear whether the effectiveness of LLM-based GER generalizes to low-resource languages.
The success of GER depends heavily on the language-specific knowledge encoded in LLMs, including lexical coverage and syntactic modeling capacity. Such knowledge may not transfer uniformly across languages, especially when pretraining data are scarce.

Another important concern when applying LLMs to ASR error correction is data contamination. Evaluation data may overlap with the text seen during LLM pretraining, potentially leading to inflated performance estimates. 
This issue is especially relevant for low-resource languages, where publicly available text is limited and often concentrated in a small number of widely used sources. As a result, performance gains observed on public datasets may partly reflect memorization rather than actual error correction ability. Therefore, a careful contamination-aware evaluation is required for assessing the real effectiveness of LLM-based GER in low-resource ASR.

\begin{figure}[t]
\centering
\includegraphics[width=\linewidth]{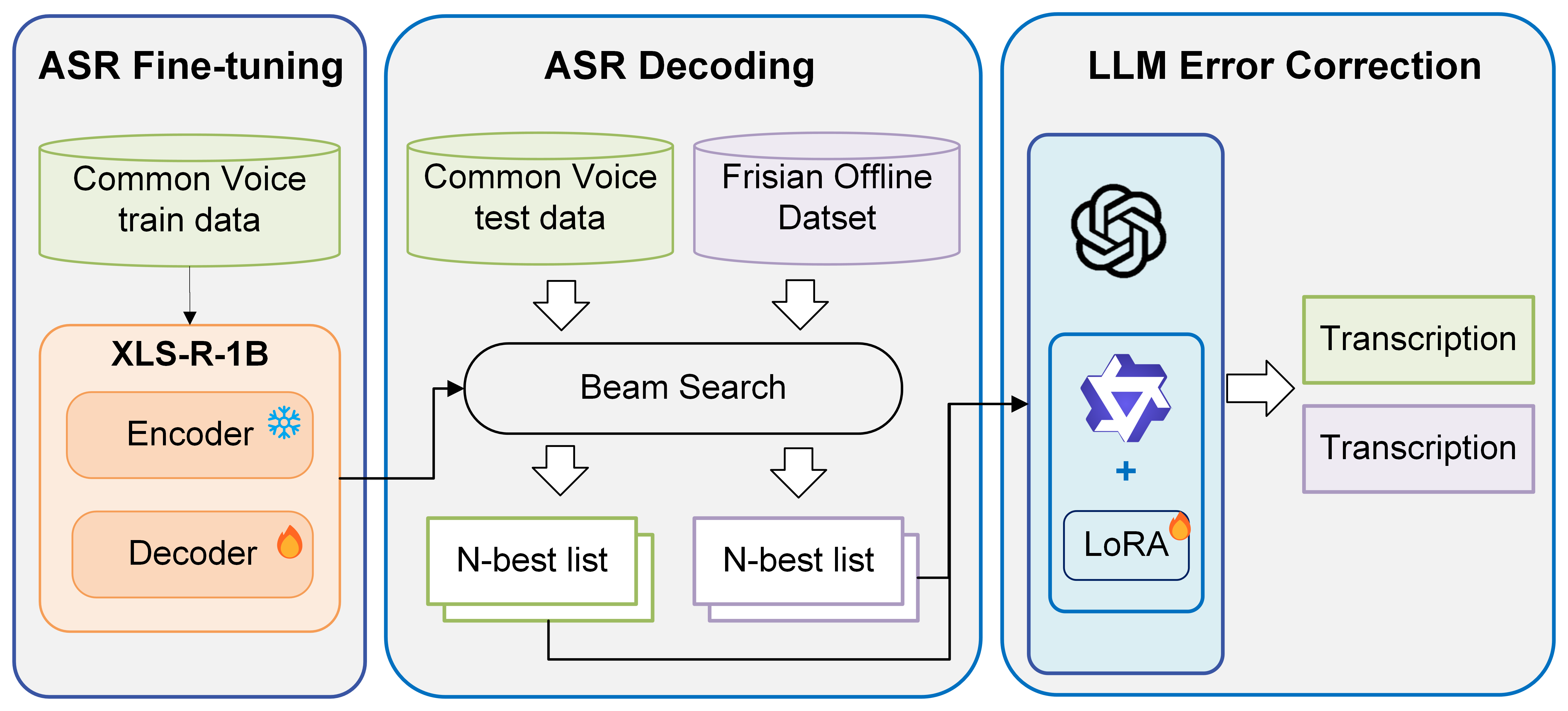}
\caption{Pipeline of the LLM-based generative error correction system. Note that we use $N = 5$ in this paper.}
\label{fig:pipeline}
\end{figure}

Motivated by these challenges, this work investigates the use of LLMs for GER in low-resource ASR, with a particular focus on West Frisian, a language spoken in the north of the Netherlands by about 400,000 speakers. In addition to the open-source Common Voice dataset~\cite{ardila2020common}, we explicitly account for data contamination by evaluating on a newly collected speech dataset with non-public text. Specifically, we address the following research questions:\footnote{The overall pipeline of our approach is shown in Figure~\ref{fig:pipeline}.} 

\vspace{0.2cm}
 \begin{itemize}
    \setlength\itemsep{1em}
    \item \textbf{RQ1:} To what extent can large language models improve ASR performance through generative error correction in a low-resource language setting?
    \item \textbf{RQ2:} Does the effectiveness of LLM-based generative error correction differ between a publicly available benchmark and a non-public dataset where data contamination is not a concern?
\end{itemize}


\section{Related Work}

\subsection{Generative Error Correction with LLMs}

Ma et al.~\cite{ma2023can} first demonstrated that generative LLMs such as ChatGPT can effectively correct ASR outputs using zero-shot and few-shot prompting with N-best hypotheses as input. Chen et al.~\cite{chen2023hyporadise} introduced HyPoradise, an open benchmark providing large-scale N-best hypotheses paired with reference transcriptions to facilitate systematic evaluation of LLM-based English GER. Yang et al.~\cite{yang2023generative} demonstrate that through instruction-based and task-activating prompting, LLMs can match domain-tuned language models in rescoring and even exceed the N-best oracle when combined with fine-tuning. Naderi et al.~\cite{naderi24_interspeech} proposed a range of confidence filtering methods to mitigate over-correction in GER.

While most early studies focused on English and other high-resource languages, recent work has begun to extend LLM-based error correction to multilingual and low-resource settings. Li et al.~\cite{li2024investigating} investigate multilingual one-best correction across 20 languages, and Yang et al.~\cite{yang2025covoger} introduce CoVoGER, a multilingual and multitask benchmark covering 15 languages for speech-to-text generative error correction. Xu et al.~\cite{xu25g_interspeech} explored using LLMs in correcting ASR errors for low-resource conversational child speech. 
However, existing studies have rarely examined the issue of data contamination in LLM-based GER for low-resource languages. In contrast to prior work, we explicitly investigate the potential impact of contamination by constructing a dataset with non-public text resources. Furthermore, we provide a detailed analysis of correction behavior across different error types to better understand how LLM-based GER improves low-resource ASR outputs.

\subsection{Data Contamination in Large Language Models}

Data contamination refers to situations in which language models have been exposed to evaluation benchmarks during training, which may lead to inflated performance estimates that do not reflect the LLM's true generalization ability~\cite{sainz-etal-2023-nlp}. This has become a central concern in the evaluation of LLMs, particularly for widely used NLP benchmarks~\cite{li2024open, deng2024investigating, chen2025benchmarking}. 
In the context of speech processing, Ma et al.~\cite{ma2025asr} examined GPT-3.5 and GPT-4 using a data contamination quiz and found measurable signs of contamination in GPT-4 based on LibriSpeech and TED-LIUM3.
However, as Balloccu et al.~\cite{balloccu2024leak} pointed out, direct contamination detection for closed-source models such as GPT-4 may be unreliable, as these systems employ mechanisms to prevent verbatim reproduction of training data, thereby masking potential memorization effects. Moreover, the likelihood of data contamination is influenced by the distribution and reuse of publicly available text. For low-resource languages, textual resources are typically scarce and drawn from a limited set of widely reused sources, which increases the probability of data contamination.

Xu et al.~\cite{xu2024benchmark} provide a systematic survey of benchmark data contamination and outline mitigation strategies, highlighting the curation of new evaluation datasets as a practical solution when pretraining data transparency is unavailable.
To better control for potential data contamination, we construct a new speech dataset with non-public textual content. This allows us to more reliably assess the generalization ability of LLM-based GER for low-resource languages.

\newpage
\section{Methods}

\subsection{Datasets}

\noindent  \textbf{Common Voice 17.0} \hspace{0.2cm}
The Common Voice corpus \cite{ardila2020common} is a massively-multilingual collection of transcribed speech intended for speech technology research and
development. The text in Common Voice primarily originates from publicly available sources, particularly Wikipedia articles, and is supplemented by community-submitted sentences that are validated by volunteers.
We adopt the official data splits as provided in the Common Voice release. The Frisian training split contains 3,921 utterances from 195 speakers (5.5 hours), whereas the validation split includes 3,170 utterances from 271 speakers (4.6 hours), and the test split comprises 3,171 utterances from 904 speakers (4.7 hours). 

\vspace{0.1cm}
\noindent \textbf{Frisian Offline Dataset} \hspace{0.2cm}
To enable contamination-aware evaluation, we construct a new speech dataset based on non-public textual prompts.\footnote{The data collection protocol was reviewed and approved by the Research Ethics Committee of our research institute.} The textual materials consist of two parts. The first part consists of 103 sentences from a Frisian storybook, for which we verified that no electronic version is available online. The second part consists of 100 original sentences written by a native Frisian speaker specifically for this study. 
Speech recordings were made in a sound-attenuated speech laboratory at our university using a head-mounted microphone. Audio was recorded in mono at a sampling rate of 44.1 kHz with 16-bit resolution. Four male native Frisian speakers participated in the recordings. 
The final dataset contains 811 utterances, totaling 1.5 hours of speech.

\subsection{ASR Model}

We adopted the XLS-R 1B model \cite{babu22_interspeech} as our ASR backbone.\footnote{In preliminary experiments, we compared several pretrained speech models, including Whisper and MMS, and found that XLS-R showed the best performance for Frisian.} XLS-R is a large-scale cross-lingual self-supervised speech model based on wav2vec 2.0 \cite{baevski2020wav2vec}, pretrained on approximately 436,000 hours of speech data across 128 languages. We fine-tuned XLS-R on the Common Voice Frisian training split for 2,000 steps using a Connectionist Temporal Classification (CTC) objective. During fine-tuning, we froze the feature extractor and updated the Transformer encoder layers. The model was trained with an effective batch size of 64, a learning rate of 5e-5, and weight decay of 5e-5.

\subsection{Large Language Models}

We used \textbf{GPT-4o-mini} and \textbf{GPT-5.1} (without reasoning) via API access as closed-source commercial models. 
We also used the open-source \textbf{Qwen3-8B} model (without reasoning) in its original pretrained form and fine-tuned on the Common Voice Frisian training split.\footnote{In preliminary experiments, we also evaluated other open-source LLMs, including Qwen2.5-7B-Instruct and Meta-Llama-3-8B-Instruct. Among the open-source LLMs, Qwen3-8B achieved the best performance and was therefore selected for further experiments.}
For fine-tuning, we generated the five best hypotheses for each training utterance using the XLS-R model. The model input consists of the zero-shot prompt together with these hypotheses, and the reference transcription serves as the training target.
We applied Low-Rank Adaptation (LoRA) to the attention and feed-forward projection layers with rank $r=16$, $\alpha=32$, and dropout $0.05$. We trained for 3 epochs with an effective batch size of 16.

\vspace{0.5cm}
\subsection{LLM-Based Generative Error Correction}

\begin{figure}[t]
\centering
\includegraphics[width=\linewidth]{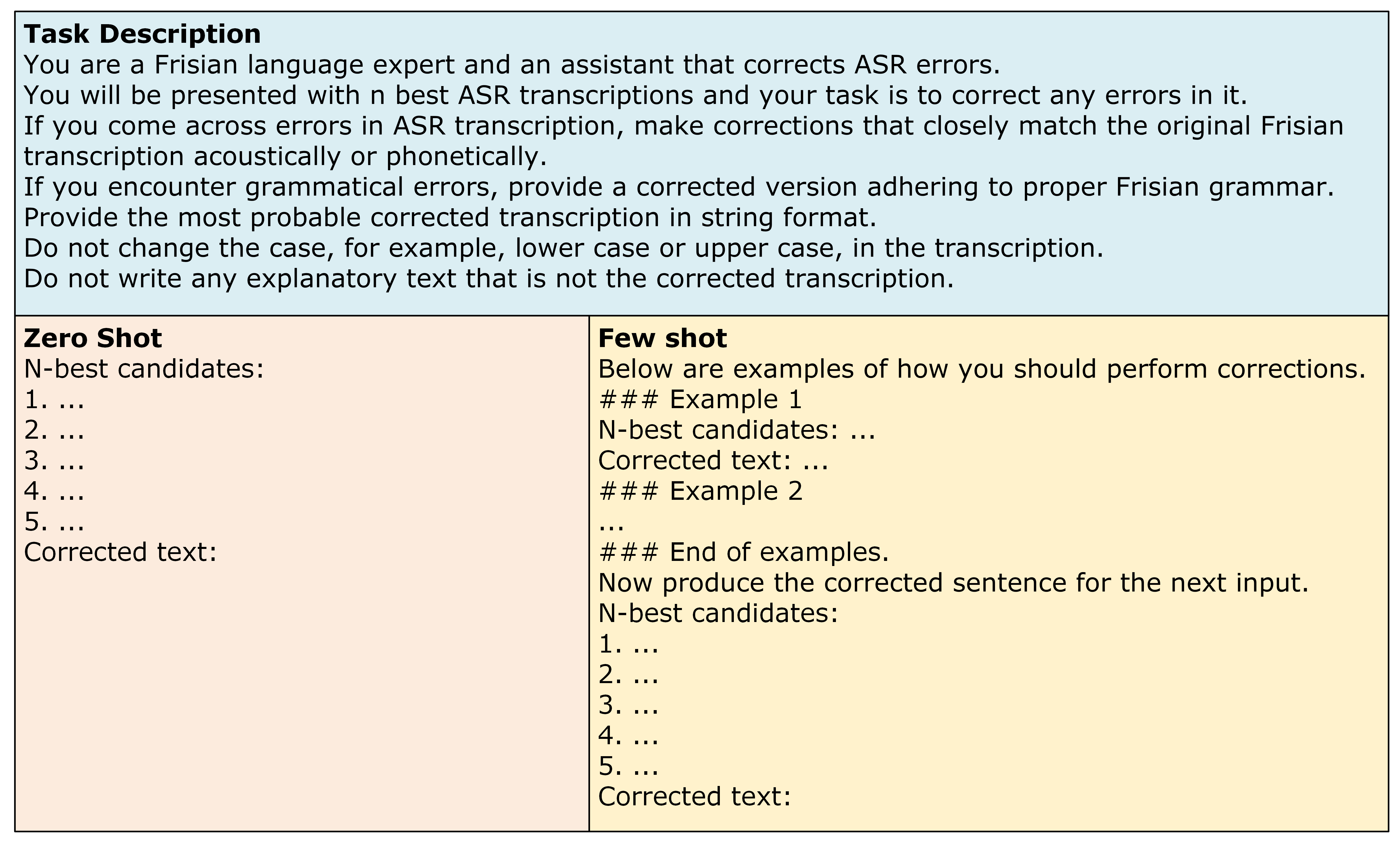}
\caption{The prompts used for our generative error correction system.}
\label{fig:prompt}
\end{figure}

For each utterance, we extract the five-best list using beam search decoding (beam width = 50) from the fine-tuned XLS-R model. These hypotheses are provided to the LLM, which is instructed to be a Frisian language expert and to provide the final corrected transcription (see Figure~\ref{fig:prompt}). In the $k$-shot ($k$=1,3,5,10) settings, we additionally prepend $k$ correction examples from the Common Voice validation split to guide the model's behavior. We also include a selection-based approach for comparison, where the model is prompted to choose the best candidate from the ASR 
five-best list and output the corresponding index.

We evaluate performance using Word Error Rate (WER), with the XLS-R one-best predictions as the ASR baseline. We also report the oracle five-best WER, defined as the minimum WER achievable by selecting the closest hypothesis to the reference from the five-best list. Furthermore, we use a trigram language model trained on the Common Voice Frisian training split into the decoding process as a conventional LM baseline. Table~\ref{tab:example} shows an example sentence of ASR error correction by different methods. It shows that the LLMs are not restricted to the N-best list, but have the freedom to correct errors outside of what was present in the best examples.

\begin{table}[t]
\vspace{0.4cm}
\centering
\footnotesize
\setlength{\tabcolsep}{2pt}
\renewcommand{\arraystretch}{1.1}
\caption{An example sentence from the Common Voice test dataset. Tokens differing from the reference are highlighted in red, and correctly corrected tokens are shown in green. The symbol (*) denotes the hypothesis selected by the selection-based GPT-5.1 method.}
\vspace{0.1cm}
\label{tab:example}
\begin{subtable}[t]{\linewidth}
\centering
\resizebox{\columnwidth}{!}{
\begin{tabular}{l p{0.85\linewidth}}
\toprule
Reference &
der wiene mar sa'n tweintich minsken op de ynformaasjej\^un \\
\midrule
Translation &
There were only about twenty people at the information evening. \\
\midrule
N-best 1 (*)&
\bad{de} \bad{wine} \bad{mat} sa'n tweintich minsken op de \bad{ynformaasje} \bad{j\^un} \\
N-best 2 &
\bad{de} \bad{wine} \bad{moat} sa'n tweintich minsken op de \bad{ynformaasje} \bad{j\^un} \\
N-best 3 &
\bad{de} wiene \bad{mat} sa'n tweintich minsken op de \bad{ynformaasje} \bad{j\^un} \\
N-best 4 &
\bad{de} wiene \bad{moat} sa'n tweintich minsken op de \bad{ynformaasje} \bad{j\^un} \\
N-best 5 &
\bad{de} \bad{wine} \bad{hat} sa'n tweintich minsken op de \bad{ynformaasje} \bad{j\^un} \\
\midrule
Trigram &
\bad{de} \good{wiene} \bad{moat} sa'n tweintich minsken op de \bad{ynformaasje} \bad{j\^un} \\
\midrule
GPT-5.1 &
\good{der} \good{wiene} \good{mar} sa'n tweintich minsken op de \good{ynformaasjej\^un} \\
\bottomrule
\end{tabular}
}
\end{subtable}
\end{table}

\begin{table}[th]
  \caption{Generative and selection-based error correction WER(\%) results for Common Voice Frisian. Zero-shot to 10-shot WERs are based on the generative method. Red numbers indicate performance worse than the XLS-R baseline, green numbers indicate performance exceeding the XLS-R-based five-best oracle, and black numbers indicate improvements over the baseline that do not surpass the oracle.}
  \vspace{-0.1cm}
  \label{tab:cv_results}
  \centering
  \setlength{\tabcolsep}{2pt}
  \resizebox{\columnwidth}{!}{
  \begin{tabular}{ p{1.7cm} rrrrrrr }
    \toprule
    \textbf{Baseline} & \textbf{WER(\%)} & & & & & & \\
    \midrule

    XLS-R           & 13.5 \\
     - oracle    & 9.6 \\
     - trigram & 12.1 \\

    \midrule\midrule

    \textbf{LLM} & \textbf{Selection} & \textbf{0-shot} & \textbf{1-shot} & \textbf{3-shot} & \textbf{5-shot} & \textbf{10-shot} \\
    \midrule

    Qwen3     & \bad{14.7} & \bad{14.4} & \bad{14.1} & \bad{13.8} & \bad{13.9} & \bad{13.9} \\
    Qwen3-FT  & 13.5 & 13.5 & 13.5 & 13.4 & 13.5 & 13.4 \\
    GPT-4o-mini      & 12.8 & 12.5 & 12.4 & 12.2 & 12.5 & 12.4 \\
    GPT-5.1      & 12.1 & 10.1 &  \good{9.5} &  \textbf{\good{8.9}} &  \good{8.9} &  \good{9.0} \\

    \bottomrule
  \end{tabular}
  }
\end{table}

\newpage

\vspace{-0.2cm}
\section{Results and Discussion}

\para{Common Voice Test Dataset}
Table~\ref{tab:cv_results} presents the WER results for the Common Voice test dataset.
We observe that, with the exception of the original (non-fine-tuned) Qwen3 model, all LLMs improve over the baseline XLS-R system. Even after LoRA fine-tuning and providing examples, Qwen3-FT yields only a marginal improvement (13.4\%), indicating limited correction capability.
Unsurprisingly, GPT-5.1 achieves the best results among all the models. In the generative setting, it substantially outperforms both the trigram baseline and the five-best oracle, reaching a minimum WER of 8.9\% under 3-shot prompting. Notably, the selection-based method achieves only 12.1\% WER, which is far from the oracle WER. This demonstrates the inherent limitation of selecting from the fixed five-best list, whereas GER is able to move beyond the hypothesis space and produce improved transcriptions not present in the original beam outputs.
Regarding few-shot prompting, moderate improvements are observed when moving from zero-shot to 1- and 3-shot settings for all the LLMs, with 3-shot achieving the best overall performance. Including more examples in the 5-shot and 10-shot settings does not appear to improve performance further.

\begin{table}[t]
  \caption{Generative and selection-based error correction WER(\%) results for Frisian Offline Data. Notations are the same as in Table~\ref{tab:cv_results}.}
  \label{tab:spraaklab_results}
  \centering
  \setlength{\tabcolsep}{2pt}
  \resizebox{\columnwidth}{!}{
\begin{tabular}{ p{1.7cm} rrrrrrr }
\toprule
\textbf{Baseline} & \textbf{WER(\%)} & & & & & & \\
\midrule

XLS-R            & 21.1 \\
 - oracle        & 18.0 \\
 - trigram       & 19.2 \\

\midrule\midrule

\textbf{LLM} & \textbf{Select} & \textbf{0-shot} & \textbf{1-shot} & \textbf{3-shot} & \textbf{5-shot} & \textbf{10-shot} & \\
\midrule

Qwen3        &   \bad{21.2}   & 21.0 & 21.0 & 20.9 & 20.8 & 20.8 \\
Qwen3-FT    & 21.0 & 20.9 & 20.9 & 20.9 & 20.9 & 20.9 \\
GPT-4o-mini  &  20.2    & 19.3 & 19.0 & 18.8 & 18.4 & 18.2 \\
GPT-5.1   &   19.9      & \good{15.3} & \good{14.7} & \good{13.9} & \good{13.8} & \textbf{\good{13.8}} \\

\bottomrule
\end{tabular}
  }
\end{table}

\para{Frisian Offline Dataset}
Table~\ref{tab:spraaklab_results} reports WER results for the Frisian offline dataset. The overall patterns are consistent with those observed for the Common Voice data. All LLMs improve over the baseline XLS-R system, except for the selection-based approach using  the original Qwen3 model. Both fine-tuned and non-fine-tuned Qwen3 variants yield only marginal improvements. Generation-based methods consistently outperform selection-based corrections. GPT-4o-mini-based GER achieves performance comparable to or better than the trigram baseline. In particular, GPT-5.1 substantially reduces WER under generative settings, reaching a minimum WER of 13.8\% (10-shot), substantially outperforming the oracle WER.
As this dataset is exclusively constructed from non-public textual prompts, the strong performance of GPT-5.1 on this dataset suggests that the observed gains cannot be explained by potential data contamination. Instead, these results provide evidence that LLM-based generative correction reflects genuine modeling capability beyond overlap with publicly available corpora, even in low-resource settings.

\vspace{-0.05cm}
\subsection{Sentence-Level Improvement}
\vspace{-0.05cm}

\begin{figure}[t]
\centering
\includegraphics[width=\linewidth]{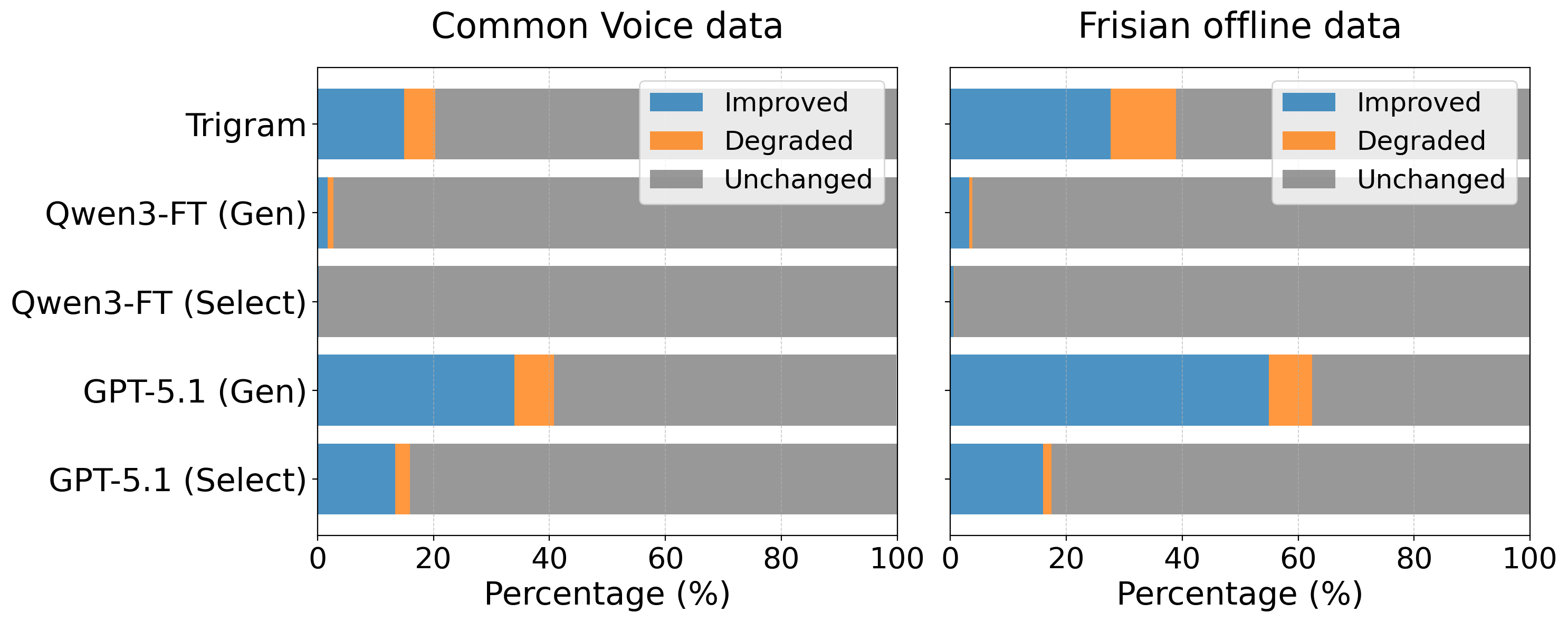}

\caption{Sentence-level improvement rates of trigram, Qwen3-FT and GPT-5.1 (generation vs.\ selection) on both Frisian ASR datasets.}
\vspace{-10pt}
\label{fig:sentence_results}
\end{figure}

To gain a better insight into how often the model genuinely improves transcriptions, we analyze sentence-level behavior of the language models by categorizing each utterance as \textit{improved}, \textit{degraded}, or \textit{unchanged} relative to the WER of the baseline ASR output. Figure~\ref{fig:sentence_results} summarizes this distribution for both datasets. For Common Voice, GPT-5.1 (Gen) improves 35.1\% of the sentences, outperforming the trigram baseline and Qwen3-FT models. However, this more aggressive correction strategy is accompanied by a relatively higher degradation rate. In contrast, GPT-5.1 (Select) improves only 13.6\% of sentences and leaves 83.7\% unchanged, reflecting the limitation of selecting within the fixed five-best list. Qwen3-FT shows minimal correction behavior in both generative and selection modes, leaving 97.3\% and 99.8\% of sentences unchanged, respectively, indicating that it largely defaults to the first hypothesis in the five-best list.

This phenomenon is even more pronounced for the Frisian offline data. GPT-5.1 (Gen) improves 54.9\% of sentences, while the degradation rate remains relatively low at 7.5\%, even below that of the trigram model (11.3\%). As the trigram model is trained on the Common Voice text, it likely suffers from domain mismatch for the Frisian offline data, while GPT-5.1 demonstrates stronger generalization across different data domains. Qwen3 again behaves in a very conservative manner for this dataset.

\vspace{-0.05cm}
\subsection{Edit-Level Error Analysis}
\vspace{-0.05cm}

We further analyze the error correction behavior of the best-performing models by examining their handling of substitutions (S), deletions (D), and insertions (I). For each error type, we compute the number of errors that are correctly adjusted by the model (true positives, TP), the number of errors that remain uncorrected (false negatives, FN), and the number of incorrect edits newly introduced by the model (false positives, FP).
Based on these counts, we compute precision and recall.

\begin{table}[t]
\centering
\scriptsize
\caption{
Edit-level correction performance of GPT-5.1 on Common Voice and Frisian offline data.
S, D, and I denote substitution, deletion, and insertion errors, respectively.
Rec. and Prec. denote recall and precision.}
\setlength{\tabcolsep}{4pt}
\resizebox{\columnwidth}{!}{
\begin{tabular}{llccccc}
\toprule
\bf Dataset / Model & \bf Type & \bf TP & \bf FN & \bf FP  &  \bf Prec. & \bf Rec. \\
\midrule
\multirow{4}{*}{\makecell{\shortstack{Common Voice\\GPT-5.1 (3-shot)}}}
& S & 1,376 & 1,765 & 281 & 83.0 & 43.8 \\
& D & 130  & 210  & 26  & 83.3 & 38.2 \\
& I & 245  & 140  & 115 & 68.1 & 63.6 \\
\cmidrule(lr){2-7}
& Total & 1,751 & 2,115 & 422 & 80.6 & 45.3 \\
\midrule
\multirow{4}{*}{\shortstack{Frisian Offline\\GPT-5.1 (10-shot)}} 
& S & 865  & 1,185 & 119 & 87.9 & 42.2 \\
& D & 124  & 228  & 14  & 89.9 & 35.2 \\
& I & 140  & 85   & 89  & 61.1 & 62.2 \\
\cmidrule(lr){2-7}
& Total & 1,129 & 1,498 & 222 & 83.6 & 43.0 \\
\bottomrule
\end{tabular}
}
\end{table}

The overall recall and precision are comparable across the two datasets, suggesting that the model exhibits consistent correction behavior regardless of whether the evaluation text is publicly available or not. Among the three error types, insertion errors show the lowest precision on both datasets with the highest recall. This suggests that the model adopts an aggressive strategy when handling extra words in ASR outputs, but it may also introduce unnecessary insertions.
In contrast, deletion errors show the lowest recall but relatively high precision. Notably, both deletion TPs and insertion FPs involve token addition: a TP for deletion indicates that the model correctly inserts a missing word, whereas an FP for insertion corresponds to the model mistakenly adding an extra word. Considering deletion TPs and insertion FPs together, the results suggest that the model struggles with token insertion decisions: sometimes adding the correct word, but at other times introducing unnecessary modifications.
Finally, substitution errors account for the majority of ASR errors, and the model exhibits intermediate performance on this error type.



\label{tab:error_analysis_combined}

\vspace{-0.05cm}
\section{Conclusion}
\vspace{-0.05cm}

This work investigated the effectiveness of LLM-based generative error correction for low-resource ASR on both public and non-public Frisian datasets. We demonstrated that GPT models can substantially improve ASR performance beyond the traditional trigram model and even surpass the five-best oracle in certain settings (RQ1). The consistent improvements observed on the non-public dataset suggest that the gains primarily stem from genuine modeling capability rather than overlap with pretraining data (RQ2). At the same time, Qwen3 achieved limited gains, together with sentence-level analysis showing that it rarely modifies the ASR outputs, indicating that effectively adapting open-source LLMs to low-resource languages remains a challenge.
Edit-level analysis reveals an asymmetric correction strategy across error types in GPT-5.1, with particular difficulty in token insertion decisions.

Future work may extend the evaluation to a wider range of LLMs and languages to better understand cross-lingual differences in correction capability. Exploring alternative input formulations and decoding strategies may further enhance GER performance. More fine-grained error analyses could also provide deeper insight into the mechanisms of LLM-based GER.







\section{Generative AI Use Disclosure}

In preparing this manuscript, we used GPT-5.1 to improve the quality of the writing, translate written text into English, and help with writing code and debugging. It was not used for writing any major part of the paper. The final content was fully reviewed by all the authors.




\bibliographystyle{IEEEtran}
\bibliography{mybib}

\end{document}